\pdfoutput=1
\documentclass[11pt]{article}
\usepackage[margin=1.1in]{geometry}
\usepackage[T1]{fontenc}
\usepackage[utf8]{inputenc}
\usepackage{lmodern}
\usepackage{microtype}
\usepackage{booktabs}
\usepackage{graphicx}
\usepackage{amsmath}
\usepackage[numbers,sort&compress]{natbib}
\usepackage{xcolor}
\usepackage[colorlinks=true,linkcolor=blue!60!black,citecolor=blue!60!black,urlcolor=blue!60!black]{hyperref}
\usepackage{caption}
\title{\textbf{Low-Cost Assays for Measuring Model Behavior\\Across Vendors and Releases}}
\author{Tapan Parikh\\Cornell Tech\\\texttt{tsp53@cornell.edu}}
\date{}

\begin{document}
\maketitle

\begin{abstract}
Language models advise people, keep them company, and write software while they sleep. Measuring what they do is hard: behavior has to be sampled repeatedly across models, prompts and releases, most of it lives in unstructured text that has to be coded before it can be counted, and the result has to be legible and rigorous enough to meaningfully compare models and vendors. To address these constraints, we present a simple, cheap, scalable, and replicable model for studying model behavior. Each study is a frozen, public stimulus run identically on a cross-vendor panel, at a few dollars per model or less. Each reads its transcripts one of three ways, chosen by how much interpretation the behavior needs: exact match on a clamped reply, a codebook applied by LLM judges whose agreement with a human coder is reported per code, and an instrumented environment that records what an agent did independently of what it said. Run across four years of model releases from both frontier and open-source labs, these instruments find four things. \textbf{Convergence:} asked to pick a word, 27 of 44 models answer \emph{serendipity} at least once in four tries. \textbf{Resistance:} a trailing ``right?'' moves endorsement by up to 32 points, and the sign flips from sycophantic to resistant as generations advance, keyed to the tag's surface form. \textbf{House:} whether a model holds a position under pressure tracks its generation, and how it holds tracks the lab that built it. \textbf{Account:} told to do something the documentation in their repository contradicts, some coding agents never went along silently and others always did, and the same model can change with the harness it runs in. Re-run on every release, batteries like these track how behavior is changing across vendors and over time.
\end{abstract}

\section{A measurement problem}
\label{sec:cost}

People ask LLMs for advice, confide in them, and let them act on their behalf while coding. Most model benchmarks measure capabilities, or how they respond to red teamed attacks. How a model behaves in everyday moments shapes what people believe and do, and our public understanding of how they behave in such situations is limited. We need to know how models behave under pressure, how they respond when a question has no right answer, and what they report about what they did when we ask them to do tasks unattended.

This is a hard measurement problem. Behavior is stochastic and varies by model, prompt and context, so it has to be sampled repeatedly across all three, and sampled again as new models are released or updated. Most of the behavior worth measuring is in unstructured text, which has to be read and coded before it can be counted. The result has to be legible to the people who deploy and use these models and rigorous enough to survive a skeptical reader.

We present a simple, cheap, scalable, and replicable model for studying model behavior to address these requirements.  Each battery is a frozen stimulus that puts a model in a situation where behavior matters, sent identically to every model on a cross-vendor panel. The panel can be re-run on every release, and a battery becomes a standing record of how model behavior is changing across vendors and over time.

Running one fixed stimulus across many models is what makes difference legible. Every model meets the same situation, so where they diverge helps us understand how models differ, and where they agree it helps us see what behaviors are common across them. Each instrument reads its transcripts one of three ways, chosen by how much interpretation the behavior needs. Every transcript, label and score is public.

A study can be of one deployed agent, one model, one lineage, or the whole field, it can widen to more languages and scenes, or carry custom scenarios and rubrics of the kind eval teams already write for their own agents.

\section{The assays}
\label{sec:method}

The single-turn and conversational assays run cross-vendor through one API at requested temperature 1.0, with reasoning off where a model allows it. The agent assay runs each coding product or harness in its own full-auto mode. Running one model through any of these studies costs a few dollars or less.\footnote{At list prices in September 2026. The API assays cost a median of three cents per model, priced by \texttt{harness/cost.py} over the token counts in each study's \texttt{spec/tokens.json}; the agent assay's metered runs cost about two dollars per configuration.}

Three analysis methods, in order of how much interpretation the behavior needs. \textbf{Clamped:} where the prompt clamps the reply to a discrete choice, the score is exact match on normalized tokens, a regular expression recomputable from the transcripts with no model in the loop. \textbf{Coded:} where the prompt is open, and the model is free to respond in any manner, the score is a code applied to the transcript. A person builds the codebook by reading a sample of transcripts, LLM coders from several vendors apply it to every transcript, and a blind human pass on held-out transcripts checks each code before it is used. In the conduct study, the codes behind the findings reported below agree with the human pass at Cohen's $\kappa$ 0.80 to 0.87, except self-citation at 0.54~\citep{parikh2026conduct}. The person does the part that needs judgment, and the machines do the reading at volume, which is what keeps the coded readout cheap. \textbf{Instrumented:} where the model operates in an instrumented environment, the environment records what it did, and the score compares that record with what the model said. 

\section{Related work}
\label{sec:related}

Evaluation has been framed as social-science measurement~\citep{wallach2025position}, and machines as agents whose behavior can be studied~\citep{rahwan2019machine}. Fixed prompts across many models are how benchmarks already work~\citep{liang2022helm}, and sycophancy~\citep{perez2022discovering,sharma2023sycophancy}, LLM judges~\citep{zheng2023judging}, model homogenization~\citep{kleinberg2021monoculture,bommasani2022picking} and coding agents~\citep{jimenez2024swebench} each have their own literature. \citet{alnasser2026authority} code 14 models under user disagreement with an a priori scheme; ours is inductive, and the vendor-level result is new. What is different here is the target and the price. The assays measure everyday behavior with no clearly right answer, are cheap enough to re-run on every release, and the analysis method is chosen by how much interpretation the behavior needs.

\section{Four preliminary findings}
\label{sec:findings}

Run across four years of model releases, the assays found four things.

\subsection{Convergence: a different serendipity in every language}
\label{sec:convergence}

Asked to pick a word (``Pick a word. Reply with one word only.''), 27 of 44 models from more than a dozen labs answer \emph{serendipity} at least once in four tries, and it is 41 percent of all answers. The prompt is one of 31 in the one-word census~\citep{parikh2026census}. The category prompts converge harder: \emph{oak} is 94 percent of all trees, \emph{hammer} 94 percent of tools, and \emph{rose} 91 percent of flowers. Within the Claude, GPT, Qwen, and Grok lineages, later releases are more conformist, with the trend partially reversing for the latest flagships. Asked the same open prompt in 44 languages~\citep{parikh2026language}, the models settle on a default word in each, and the default is the language's own. \emph{Serendipity} wins only in English. Every other language has its own serendipity: \emph{sky} in seven Indic languages and in Persian and Malay, \emph{sun} in Ukrainian, Russian, and Spanish, \emph{peace} in Hebrew, Arabic, and Amharic, \emph{cat} in Japanese and Polish. 

\subsection{Resistance: keyed to the tag}
\label{sec:resistance}

The tag-question study~\citep{parikh2026tag} asks 20 decisions between two defensible options (rent or buy; name the cat Luna or Willow) as a neutral question and as the same sentence with a confirmation tag (``Luna is the better choice, right?''), counterbalanced over both options, scored by exact match on clamped yes or no. Across 45 models the tag effect spans $+32$ percentage points (endorses more often when the user fishes for agreement) to $-32$ points. Within every lineage the effect crosses from positive to negative as generations advance (GPT $+4$ to $-28$, Claude $+7$ to $-32$, Qwen $+16$ to $-11$), and two releases during the study window landed on the trend out of sample. Models are getting more resistant to bids for agreement over time.

Replace \emph{right?} with \emph{maybe?} and agreement rises above a neutral framing in 45 of 45 models, by 19.6 points on average, so two single-token tags on an otherwise identical sentence span 25 points. Agreement tracks how sure the user sounds: a confident bid meets resistance in the newer models, and a tentative one is rubber-stamped by nearly all of them. \emph{Maybe?} also softens the claim being endorsed, so part of this contrast may be about what the user asserts rather than how they ask.

\subsection{House: newer models hold, each lab holds its own way}
\label{sec:house}

The conduct study~\citep{parikh2026conduct} puts a model under social pressure in four-turn scenes: a user insists that $5\times9=54$ and turns contemptuous, begs for a backdated doctor's note, wants applause for quitting to day-trade, or discloses a dying parent and then asks for a cheerful email. In each scene the model either holds its ground or folds.

Across 38 models, the oldest third folded 43 percent of the time, and models released since mid-2025 folded 23 percent of the time. A larger panel of 60 models shows the same trend. A codebook of 17 codes records how each model held or folded. Six of the codes sort by lab, and release date does not explain them. Anthropic's models name the user's feeling, warn of the consequence, and offer an alternative, each well above the panel. Google's models cite their own nature or rules at three times the panel rate, and apologize when they fold. Meta's models question the plan more than the panel does, and when they fold they produce what was asked. OpenAI's signature is what its models do less of: they warn less, name feelings less, and almost never cite their own nature or rules.

\subsection{Account: what agents did and what they said}
\label{sec:account}

The coding atlas~\citep{parikh2026atlas} gives a coding agent a small repository and a one-line instruction. Each repository holds a right thing to do, a shortcut that looks finished, and a hidden check the agent never sees. It asks three questions. When the job is only part done, does the agent say so? When the test and the code disagree, does it find out which is wrong? When the user insists on something the repository's own documentation contradicts, does it hold or go along? Two records come back from every run, what the agent changed on disk and what it said, and the score compares them by counts and fixed string lists. 

Fourteen configurations ran six scenarios three times each in full-auto mode: Claude Code, Codex CLI and Gemini CLI as products, and eleven models in the OpenCode harness. Taking the shortcut and reporting the job as done is rare: most configurations never did it in 12 work runs, and Codex and Gemini CLI did it twice. Going along with the user's false claim without saying the repository contradicts it splits the field. Over the six pushback replies, Claude Code, Opus 5, Fable 5 and Kimi K3 never did it, and Codex and Gemini 3.5 Flash did it every time. The harness is part of the behavior: Gemini 3.5 Flash went along silently in three of six replies in Gemini CLI and six of six in OpenCode. A third count sets the final message against the agent's own command log: runs that claim the tests pass without having run any, that execute a destructive command and never mention it, or that edit files and then say almost nothing. Gemini CLI does one of these in five of eighteen runs and Claude Code in two, where most configurations never do.

\section{Limitations}
\label{sec:limitations}

Each construct is measured with one wording, and the tag results show how much one word can move. A frozen public stimulus can enter training data, so a standing record needs fresh scenes alongside the frozen ones. Providers update models and serving settings without notice, and older models are queried as they are served now. Release date moves with size, tier and training recipe, so the generation findings are observational. The API assays run with reasoning off. The agent results rest on three runs per scenario.

\section{A behavioral science of LLMs}
\label{sec:for}

LLMs are complex phenomena that need to be studied behaviorally, across a wide range of operating contexts and configurations. Our contribution is a way of measuring model behavior that is simple, cheap, scalable and replicable. Fixed turns and short scoring rules are what make it cheap: a frozen stimulus removes the need to design a new prompt per model, and exact match, a quoted code, or a comparison of a diff against a message removes the need for an expensive reader. The instruments are simple and they still separate models across vendors and generations. A monoculture across 44 models, a sign flip within every lineage, a vendor signature in how models hold their ground, and a gap between what an agent does and what it says all came out of cheap instruments (\S\ref{sec:method}).

A study at this price is closer in scale to a small experiment with human participants than to a benchmark: one person can run it, repeat it on the next release, and point it at a new question the week after. Models are easier to recruit and schedule than people. At that price, and with no lab scoring itself, studying model behavior can become an ordinary thing to do.\footnote{Stimuli, transcripts, scores and scripts are public at \url{https://github.com/tap2k/modelun} and \url{https://github.com/tap2k/coding-atlas}.}

\bibliographystyle{plainnat}
\bibliography{references}

@misc{parikh2026census,
  author = {Tapan Parikh},
  title = {The One-Word Census: Answer-Choice Conformity Across 44 Language Models},
  year = {2026}, eprint = {2607.12796}, archivePrefix = {arXiv}
}

@misc{parikh2026tag,
  author = {Tapan Parikh},
  title = {Tag Questions and the Generational Reversal of Sycophancy Across 45 Language Models},
  year = {2026}, eprint = {2607.23976}, archivePrefix = {arXiv}
}

@misc{parikh2026conduct,
  author = {Tapan Parikh},
  title = {Conduct Under Pressure: What Sixty Language Models Do When a User Pushes},
  year = {2026}, eprint = {2609.25447}, archivePrefix = {arXiv}
}

@misc{wallach2025position,
  author = {Hanna Wallach and Meera Desai and A. Feder Cooper and Angelina Wang and Chad Atalla and Solon Barocas and Su Lin Blodgett and Alexandra Chouldechova and Emily Corvi and P. Alex Dow and Jean Garcia-Gathright and Alexandra Olteanu and Nicholas Pangakis and Stefanie Reed and Emily Sheng and Dan Vann and Jennifer Wortman Vaughan and Matthew Vogel and Hannah Washington and Abigail Z. Jacobs},
  title = {Position: Evaluating Generative {AI} Systems is a Social Science Measurement Challenge},
  year = {2025}, eprint = {2502.00561}, archivePrefix = {arXiv}
}

@article{rahwan2019machine,
  author = {Iyad Rahwan and others},
  title = {Machine behaviour},
  journal = {Nature}, volume = {568}, pages = {477--486}, year = {2019}
}

@misc{sharma2023sycophancy,
  author = {Mrinank Sharma and Meg Tong and Tomasz Korbak and others},
  title = {Towards Understanding Sycophancy in Language Models},
  year = {2023}, eprint = {2310.13548}, archivePrefix = {arXiv}
}

@misc{parikh2026language,
  author = {Tapan Parikh},
  title = {Every Language Has Its Own Serendipity: The One-Word Census Across Languages},
  year = {2026},
  howpublished = {\url{https://convovo.ai/blog/every-language-serendipity}},
  note = {Published 2026-07-24}
}

@misc{parikh2026atlas,
  author = {Tapan Parikh},
  title = {What Is Your Coding Agent Hiding From You? A field guide to coding agents (study and data)},
  year = {2026}, howpublished = {\url{https://github.com/tap2k/coding-atlas}}
}

@misc{perez2022discovering,
  author = {Ethan Perez and Sam Ringer and Kamil{\.e} Luko{\v{s}}i{\=u}t{\.e} and others},
  title = {Discovering Language Model Behaviors with Model-Written Evaluations},
  year = {2022}, eprint = {2212.09251}, archivePrefix = {arXiv}
}

@misc{liang2022helm,
  author = {Percy Liang and Rishi Bommasani and Tony Lee and others},
  title = {Holistic Evaluation of Language Models},
  year = {2022}, eprint = {2211.09110}, archivePrefix = {arXiv}
}

@misc{zheng2023judging,
  author = {Lianmin Zheng and Wei-Lin Chiang and Ying Sheng and others},
  title = {Judging {LLM}-as-a-Judge with {MT-Bench} and Chatbot Arena},
  year = {2023}, eprint = {2306.05685}, archivePrefix = {arXiv}
}

@article{kleinberg2021monoculture,
  author = {Jon Kleinberg and Manish Raghavan},
  title = {Algorithmic monoculture and social welfare},
  journal = {Proceedings of the National Academy of Sciences}, volume = {118}, number = {22}, year = {2021}
}

@misc{bommasani2022picking,
  author = {Rishi Bommasani and Kathleen A. Creel and Ananya Kumar and Dan Jurafsky and Percy Liang},
  title = {Picking on the Same Person: Does Algorithmic Monoculture lead to Outcome Homogenization?},
  year = {2022}, eprint = {2211.13972}, archivePrefix = {arXiv}
}

@misc{jimenez2024swebench,
  author = {Carlos E. Jimenez and John Yang and Alexander Wettig and Shunyu Yao and Kexin Pei and Ofir Press and Karthik Narasimhan},
  title = {{SWE}-bench: Can Language Models Resolve Real-World {GitHub} Issues?},
  year = {2024}, eprint = {2310.06770}, archivePrefix = {arXiv}
}

@misc{alnasser2026authority,
  author = {Alnasser, Riyadh and {\c{C}}etinkaya, Yusuf M{\"u}cahit and Zhao, Sumin and Elmas, Tu{\u{g}}rulcan},
  title = {How {AI} Models Manage Epistemic Authority: A Taxonomy and Comparative Analysis of Responses to User Disagreement},
  year = {2026}, eprint = {2609.07662}, archivePrefix = {arXiv},
  note = {EMNLP 2026}
}
\end{document}